\documentclass[lettersize,journal]{IEEEtran}
\usepackage{amsmath,amssymb,amsfonts}
\usepackage[scr=boondox]{mathalfa}
\usepackage{array}
\usepackage{textcomp}
\usepackage{stfloats}

\usepackage{url}
\usepackage{bm}
\usepackage{xcolor}
\usepackage{color}
\usepackage{lscape}
\usepackage{multirow}
\usepackage{array}
\usepackage{subcaption}
\usepackage{booktabs}
\usepackage{verbatim}
\usepackage{graphicx}
\usepackage{booktabs}
\usepackage[ruled,linesnumbered]{algorithm2e}
\usepackage{algorithmic}

\def\BibTeX{{\rm B\kern-.05em{\sc i\kern-.025em b}\kern-.08em
    T\kern-.1667em\lower.7ex\hbox{E}\kern-.125emX}}
\usepackage[hidelinks]{hyperref} 
\begin{document}
\title{HGFormer: A Hierarchical Graph Transformer Framework for Two-Stage Colonel Blotto Games via Reinforcement Learning}

\author{Yang Lv, Jinlong Lei 

\thanks{Manuscript created November, 2023; The paper was sponsored by the National Key Research and Development Program of China under No 2022YFA1004701,  the National Natural Science Foundation of China under Grant  No. 72271187 and No. 62373283, and partially by Shanghai Municipal Science and Technology Major Project No. 2021SHZDZX0100. 

Yang Lv is with Shanghai Research Institute for  Intelligent Autonomous Systems,  Tongji University Shanghai 200092, China, (email: 726564418@qq.com).

Jinlong Lei is with the Department of Control Science and Engineering, Tongii University, Shanghai, 201804, China; The Shanghai Research Institute for Intelligent Autonomous Systems,Shanghai, 201804, China;Shanghai Institute of Intelligent Science and Technology, Tongji University 200092, China, (email: leijinlong@tongji.edu.cn; pengyi@tongji.edu.cn).

This article has supplementary material provided by the authors available
at}
}

\markboth{IEEE TRANSACTIONS ON CYBERNETICS,~Vol.~, No.~, ~}%
{HGFormer: A Hierarchical Graph Transformer Framework for Two-Stage Colonel Blotto Games via Reinforcement Learning}

\maketitle
\begin{abstract}
Two-stage Colonel Blotto game models a typical adversarial resource allocation problem, in which two opposing agents sequentially allocate resources in a network topology across two phases: an initial resource deployment followed by multiple rounds of dynamic reallocation adjustments. The sequential dependency between game stages and the complex constraints imposed by the graph topology  make it difficult for traditional approaches to attain a globally optimal strategy. To address these challenges, we propose a hierarchical graph Transformer framework called HGformer. By incorporating an enhanced graph Transformer encoder with structural biases and a two-agent hierarchical decision model, our approach enables efficient policy generation in large-scale adversarial environments. Moreover, we design a layer-by-layer feedback reinforcement learning algorithm that feeds the long-term returns from lower-level decisions back into the optimization of the higher-level strategy, thus bridging the coordination gap between the two decision-making stages. Experimental results demonstrate that, compared to existing hierarchical decision-making or graph neural network methods, HGformer significantly improves resource allocation efficiency and adversarial payoff, achieving superior overall performance in complex dynamic game scenarios. 
\end{abstract}


\section{Introduction}\label{section1}
\IEEEPARstart{T}{he}   Colonel Blotto game has attracted extensive attention as a classic model of adversarial resource allocation,  see e.g., \cite{2,1,3}. Its  standard model assumes that two opposing players simultaneously allocate their limited resources across multiple independent resource nodes, and the winner in each resource node is the one who commits more resources. This model captures the strategic balance problem under resource constraints and has significant theoretical value and application prospects \cite{4,5}.

However, in real-world scenarios, resource allocation confrontations are often not one-shot static engagements, but instead exhibit multi-stage dynamics and spatial constraints \cite{6,7,8}. For example, at competitive strategic points, one agent (Blue side) first distributes resources based on strategic significance. After observing this distribution, the opposing agent (Red side) strategically reallocates its resources to target critical areas; subsequently, both sides can continue to maneuver and reallocate resources over multiple rounds of confrontation to respond to resource node developments \cite{9,10}. Similarly, in a network security scenario, the first decision-maker allocates protective resources to critical nodes. After observing this configuration, the second decision-maker selects target nodes to compete for control. Both parties then iteratively adjust their resource allocation across the network in a sustained strategic process \cite{11,12,13}.   Such two-stage dynamic Colonel Blotto games are far more complex than the classic model: the decision process is sequential and hierarchical, strategy choices at different times are strongly coupled, and they are constrained by the underlying network topology in terms of movement costs and connectivity.
 
The central challenge of this problem lies in coordinating strategies across multiple stages within a graph-based environment to maximize cumulative adversarial payoff. First, decisions in the initial resource allocation stage significantly influence the subsequent dynamic interactions. During this phase, one agent seeks to allocate resources across nodes in a balanced fashion to gain long-term advantages, while the other aims to preemptively secure control over strategically critical nodes. A major difficulty is accurately evaluating each node's strategic importance while accounting for global topological features of the network \cite{14}.
Second, in the dynamic reallocation stage, both agents must adapt their strategies in each round based on the evolving resource distribution and contested nodes. This phase naturally formulates a multi-step Markov decision process involving a high-dimensional state space and a continuous action space, both shaped by the underlying graph structure, which results in a vast and complex strategy search space \cite{15}. One side must reinforce its previously held positions to prevent being overrun, while also reallocating resources to exploit emerging weaknesses. Conversely, the opponent must efficiently redistribute its resources to maintain control over multiple threatened areas. Strategic decisions at this stage must carefully balance short-term tactical gains with long-term strategic outcomes.
Finally, the interdependence between the two stages introduces significant challenges in credit assignment. Initial deployment strategies that yield short-term benefits may compromise long-term objectives, and vice versa. Without an effective mechanism to integrate feedback from the dynamic stage into refining the initial allocation policy, the two layers of decision-making risk operating in isolation, thereby failing to achieve globally optimal performance.

Existing works have significant limitations in addressing the above challenges simultaneously. From the perspective of graph-based games, traditional graph algorithms or heuristics \cite{17,26,27,29} (for example, centrality-based metrics or greedy strategies) cannot fully consider the evolving interplay of adversarial strategies or capture long-range dependencies, often resulting in suboptimal static allocations. Graph neural network (GNN) methods \cite{21,22} can indeed extract features from graph-structured data, but common models based on local neighborhood aggregation struggle with long-distance dependencies and fail to capture the global state of a complex resource node network. In terms of two-stage strategy optimization, a common approach is to treat the initial allocation and the dynamic reallocation as separate subproblems solved in isolation \cite{30,31}. However, this hierarchical yet disjoint training strategy leads to a lack of coordination: the upper-level (initial deployment) policy does not account for the long-term outcomes of the lower-level dynamic game, and the effectiveness of the lower-level policy is constrained by a fixed upper-level decision, limiting the overall performance. Some recent reinforcement learning methods have been attempted for complex game decision-making (see e.g., \cite{33}), but directly applying a single-agent deep reinforcement learning algorithm in a large-scale graph-based adversarial environment often fails to converge or yields unstable strategy performance. Moreover, in the   context of two-stage dynamic Colonel Blotto games, there is  a lack of a mature end-to-end learning framework that simultaneously handles graph structural information, hierarchical decision processes, and multi-agent adversarial interactions.

To tackle these challenges, we propose a hierarchical decision-making framework for two-stage dynamic games on graphs, called HGformer. The core contributions of our work are summarized as follows:

1) Enhanced Graph Transformer Encoder (EGTE): We design a Transformer-based graph encoder that integrates a shortest-path distance bias and a dynamic virtual node to capture both long-range dependencies and global resource node context. This allows each node’s representation to reflect local structure and global strategy dynamics.

2) Hierarchical Cooperative Decision Architecture: Built on EGTE, our framework introduces a Planner Agent for initial allocation and a Transfer Agent for dynamic reallocation. The Planner uses a Pointer Network with cross-attention to iteratively select key nodes. The Transfer Agent employs a GAT-based dual-scale encoder to generate real-time, context-aware transfer strategies. Together, they enable precise, adaptive decision-making under complex constraints.

3) Layer-by-Layer Feedback Reinforcement Training (LFRT): We introduce a hierarchical joint training scheme. The Planner is first trained independently; then, as the Transfer Agent learns via PPO, its cumulative rewards are fed back to refine the Planner’s policy via REINFORCE. This feedback loop enables end-to-end coordination and optimizes global performance.

The remainder of this paper is organized as follows. Section II introduces the two-stage dynamic Colonel Blotto game model on a graph, including the formal problem formulation and underlying assumptions. Section III describes the proposed HGformer framework in detail. Section IV presents the experimental setup and result analysis, showcasing the effectiveness of our method across a range of complex adversarial scenarios. Finally, Section V concludes the paper and outlines potential directions for future research.

\section{The Two-stage Dynamic Colonel Blotto game }\label{section3}
In this section, we formally describe the structure and dynamics of the two-stage dynamic Colonel Blotto game on a graph and establish its mathematical formulation.

\subsection{Game Description }\label{section3.1}
We study a two-stage Colonel Blotto game on a graph, where nodes represent resource sites and edges denote feasible transfer paths. Red and Blue possess finite resource endowments to allocate and subsequently redistribute. A node is controlled by the side with the higher allocation; ties favor Red. Winning resources persist, while losing ones are eliminated, capturing irreversible local losses.

\begin{figure}[h]
\centering
\includegraphics[width=0.5\textwidth]{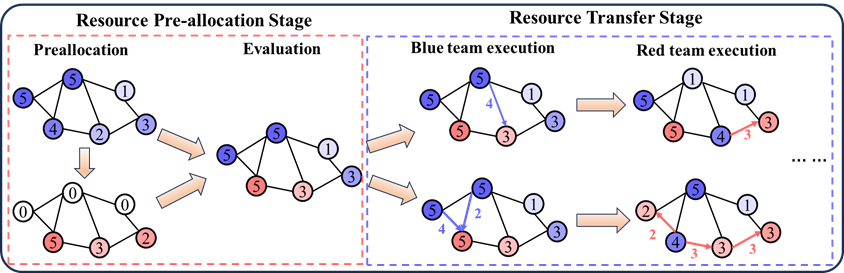}
\caption{Illustration of the two-stage Colonel Blotto game.}
\label{fig 1}
\end{figure}

As shown in Fig.~\ref{fig 1}, the confrontation unfolds in two stages:
(1) Initial Pre-Allocation Stage: Blue first distributes its resources across the graph. Observing this static configuration, Red then allocates its resources. This sequential move order gives Red perfect information but no chance to revise its placement.
(2) Dynamic Transfer Stage: Blue and Red iteratively reallocate remaining resources along the graph edges. Blue reinforces vulnerable nodes, while Red responds adaptively, probing for weaknesses. This process continues until resources are depleted or a fixed round limit is reached.

Each side seeks to maximize the total value of its controlled nodes. The coupling of global placement and local, time-sensitive maneuvers makes the game a combinatorial and sequential optimization challenge.

\subsection{Problem Formulation }\label{section3.2}
We formally model the Colonel Blotto game environment as a weighted, connected, undirected graph $\mathcal{G} = (\mathcal{N}, \mathcal{E}, \mathcal{W})$, where $\mathcal{N}$ is the set of nodes (resource locations), and $\mathcal{E}$ is the set of edges. An edge $(i,j)\in \mathcal{E}$ exists if and only if nodes $i$ and $j$ are directly connected, permitting resource transfer between them. The set of edge weights is denoted by $\mathcal{W}=\{w_{ij}\}_{(i,j)\in \mathcal{E}}$, where $w_{ij}$ signifies the distance or cost associated with transferring resources between nodes $i$ and $j$. Each node $i\in\mathcal{N}$ has a strategic value denoted by $v_i$, commonly recognized by both sides. Furthermore, let $\mathcal{N}_i$ represent the set of nodes directly connected to node $i$, i.e., $\mathcal{N}_i={j|(i,j)\in \mathcal{E}}$. Thus, the graph topology facilitates resource mobility across adjacent nodes, with transfer efficiency constrained by edge weights.

{\bf Players’ resources and distribution}: Let $\mathcal{R}^{b}$ be Blue’s initial total resource, and $\mathcal{R}^{r}$ be Red’s initial total resource, with $\mathcal{R}^{r} < \mathcal{R}^{b}$ (Red has slightly fewer resources, reflecting a defensive advantage for Blue). Throughout the game, both players must obey a resource conservation constraint: at any time, the total resources they have deployed across all nodes cannot exceed their remaining available resources. Use $\mathscr {s}_{i}^{b}(t)$ and $\mathscr {s}_{i}^{r}(t)$ to denote the resources held by Blue and Red at node $i$ at the start of round $t$ (i.e., before any conflict occurs).

{\bf Initial allocation stage}: The initial stage is modeled as a one-shot allocation game at time $t = 0$. In this phase, Blue first distributes its total resources $\mathcal{R}^{b}$ across the nodes, resulting in an initial allocation vector $\mathcal{S}_{0}^{b} = {\mathscr{s}_{1}^{b}(0), \mathscr{s}_{2}^{b}(0), \dots, \mathscr{s}_{N}^{b}(0)}$, typically satisfying $\sum_{i \in \mathcal{N}} \mathscr{s}_{i}^{b}(0) = \mathcal{R}^{b}$. After observing Blue's deployment, Red then allocates its resources, forming its own initial allocation vector $\mathcal{S}_{0}^{r} = {\mathscr{s}_{1}^{r}(0), \mathscr{s}_{2}^{r}(0), \dots, \mathscr{s}_{N}^{r}(0)}$, with $\sum_{i \in \mathcal{N}} \mathscr{s}_{i}^{r}(0) = \mathcal{R}^{r}$. Upon completion of the initial allocations, a first resolution is immediately conducted based on the above rules to determine the control and remaining resources at each node—resources at a node are reset to zero for the losing side. The resulting resource quantities, $\mathscr{s}_{i}^{r}(1)$ and $\mathscr{s}_{i}^{b}(1)$, serve as the initial state for the dynamic phase that follows.

{\bf Dynamic resource transfer stage}: In each subsequent game round $t = 1, 2, \dots$, both Red and Blue first execute their respective resource transfer strategies, followed by a resolution phase to determine the outcome of local competitions.

Formally, Red’s and Blue’s transfer strategies at round \$t\$ are defined as proportion matrices $\mathcal{U}_{t}^{r}=\{\mu_{i\rightarrow j}^{r}(t)\mid i\in\mathcal{N},j\in\mathcal{N}_{i}\}$ and $\mathcal{U}_{t}^{b}=\{\mu_{i\rightarrow j}^{b}(t)\mid i\in\mathcal{N},j\in\mathcal{N}_{i}\}$, where $\mu_{i\rightarrow j}^{r}(t)$ represents the proportion of Red’s resource at node $i$ that Red will transfer from node $i$ to a neighboring node $j$ during round $t$. If $j$ is not a neighbor of $i$ ($j \notin \mathcal{N}_{i}$), then $\mu_{i\rightarrow j}^{r}(t)$  is not applicable (we consider it effectively 0). Based on this proportional strategy, the actual amount of resource moved from node $i$ to node $j$ by Red in that round is:
\begin{equation}
    \Delta \mathscr {s}_{i\to j}^{r}(t)=\mu_{i\to j}^{r}(t)\mathscr {s}_{i}^{r}(t).
\end{equation}
To ensure feasibility, each node cannot send out more resource than it currently has. Thus, for any node $i$: $\sum_{j \in \mathcal{N}(i)} \mu_{i\to j}^{r}(t) \le 1$. The portion $1 - \sum_{j \in \mathcal{N}(i)} \mu_{i\to j}^{r}(t)$ (if any) is the fraction of resource retained at node $i$. Blue’s transfer strategy $\mathcal{U}_{t}^{b}$ satisfies a corresponding constraint for each node.

After the transfer actions are executed, the temporary resource distribution at each node is updated. For Red at node $i$, the temporary resource amount at the end of round $t$ is given by:
\begin{equation}
\tilde{\mathscr s}_i^r(t)=\mathscr {s}_{i}^{r}(t)-\sum_{j\in\mathcal{N}_{i}}\Delta \mathscr {s}_{i\to j}^{r}(t)+\sum_{k\in\mathcal{N}_{i}}\Delta \mathscr {s}_{k\to i}^{r}(t),\label{trans}
\end{equation}
where the first term subtracts the resources sent out from $i$, and the second term adds resources received from each neighbor $k$ that sent resources into $i$. Blue’s node resources $\tilde{\mathscr s}_i^b(t)$ update similarly. 

Next, the competitive outcome at each node is determined. For each node $i$, if $\tilde{\mathscr{s}}_i^r(t) \ge \tilde{\mathscr{s}}_i^b(t)$, Red captures or retains control, with updated states $\mathscr{s}_{i}^{r}(t+1) = \tilde{\mathscr{s}}_i^r(t)$ and $\mathscr{s}_{i}^{b}(t+1) = 0$. Otherwise, Blue prevails, and the updates are $\mathscr{s}_{i}^{b}(t+1) = \tilde{\mathscr{s}}_i^b(t)$ and $\mathscr{s}_{i}^{r}(t+1) = 0$.

In this resource redistribution, transfers are not free: each transfer along an edge incurs a cost proportional to the edge weight and the amount moved. Assuming a linear dissipation model, the cost of transferring resources from node $i$ to node $j$ can be defined as
\begin{equation}
    C^{r}_{i\to j}=w_{ij}\Delta \mathscr {s}_{i\to j}^{r}(t),
\end{equation}
i.e. proportional to the distance $w_{ij}$ and the amount moved. The total transfer cost incurred by Red in round $t$ is then 
\begin{equation}
    C^{r}(t)=\sum_{(i,i)\in\mathcal{E}}w_{ij}\Delta \mathscr {s}_{i\rightarrow j}^{r}(t)
\end{equation}
and Blue’s cost $ C^{b}(t)$ is defined analogously.

{\bf Strategy space and utility}: Red’s overall strategy consists of an initial allocation strategy and a sequence of dynamic transfer strategies, denoted $\pi^{r}=\{\mathcal{S}_{0}^{r},\mathcal{U}_{1}^{r},\mathcal{U}_{2}^{r},\ldots\}$. Here $\mathcal{S}_{0}^{r}$ is a choice from all possible initial deployment schemes, and each $\mathcal{U}_{t}^{r}$ is chosen from all strategies satisfying the above constraints. Blue’s overall strategy is defined similarly. The game ends when either Red or Blue runs out of resources, or a maximum number of rounds is reached. At the end of the game, based on final node ownership and cumulative resource consumption, we define each side’s utility (total payoff). Assume that at the end of the game (time $\mathcal{T}$), node $i$ is controlled by Red; then Red gains a payoff equal to that node’s value $v_i$. If node $i$ is still controlled by Blue, then Red’s payoff from that node is 0 and Blue gains $v_i$. Additionally, each side incurs a cost equal to the total resource transfer cost they spent over the game, which is deducted from their payoff. Formally, Red’s utility can be expressed as
\begin{equation}
    U_{r}=\sum_{i=1}^{N}\mathbb{I}(\mathscr {s}_{i}^{r}(\mathcal{T})\geq \mathscr {s}_{i}^{b}(\mathcal{T}))\cdot v_{i}-\sum_{t=0}^{\mathcal{T}}C^{r}(t),
\end{equation}
where $\mathbb{I}{(\cdot)}$ is an indicator (1 if Red successfully controls node $i$ at the end, 0 otherwise). Blue’s utility is the total value of nodes Blue holds minus Blue’s cumulative transfer costs: 
\begin{equation}
    U_{b}=\sum_{i=1}^{N}\mathbb{I}(\mathscr {s}_{i}^{b}(\mathcal{T})\geq \mathscr {s}_{i}^{b}(\mathcal{T}))\cdot v_{i}-\sum_{t=0}^{\mathcal{T}}C^{b}(t),
\end{equation}

Throughout the initial and dynamic stages, Red seeks to maximize $ U_{r}$, while Blue seeks to maximize $ U_{b}$. An equilibrium strategy profile for the two-stage game must balance the trade-off between node capture rewards and resource expenditure costs across both stages.

\section{Hierarchical Graph Transformer (HGformer)}\label{subsection3}
In the two-stage Blotto game, Red must overcome Blue’s larger initial endowment and its subsequent defensive maneuvers.  We address this challenge with **HGformer**, a hierarchical \emph{Planning–Transfer} architecture underpinned by an Enhanced Graph Transformer Encoder (EGTE) that yields rich, topology-aware state embeddings.

\textbf{Planner Agent (upper layer):}  Conditioned on Blue’s opening deployment and EGTE’s global embedding, the Planner emits a one-shot allocation that balances early territorial gain against long-term flexibility.

\textbf{Transfer Agent (lower layer):}  Across dynamic rounds the Transfer Agent observes node ownership, residual resources, and Blue’s reactions, and selects a full transfer matrix to maximize discounted utility.

Algorithm~\ref{alg1} executes as follows:  
1) the Planner produces Red’s initial placement;  
2) each round, the Transfer Agent generates transfers from the current global state;  
3) Blue replies via a fixed rule (proportional reinforcement of threatened or captured nodes);  
4) both sides update resources and node control until resources are exhausted or the round limit is reached.  

\begin{figure*}[h]
  \centering
  \includegraphics[width=0.95\textwidth]{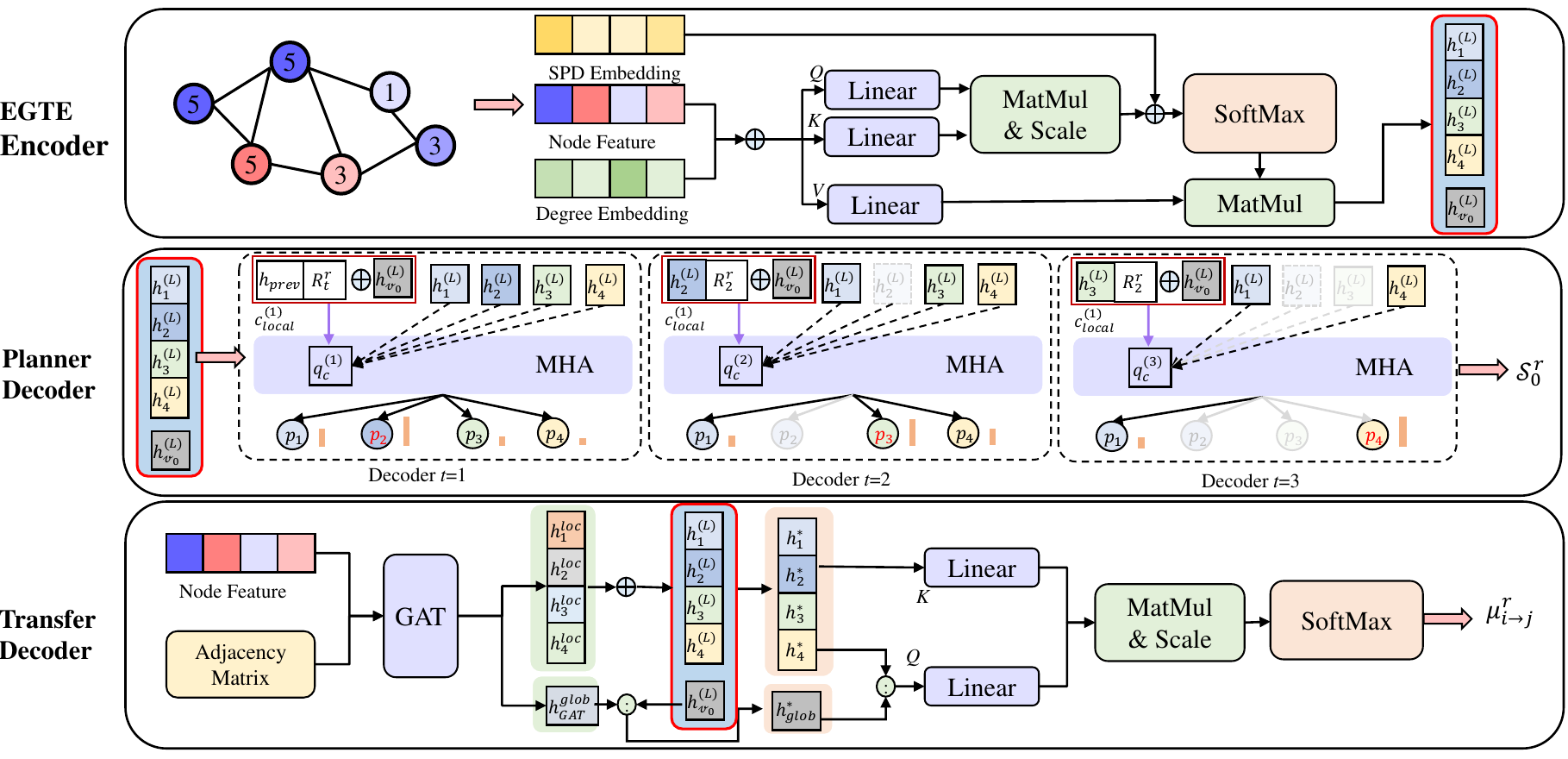}
  \caption{illustrates the detailed flow of the node resource allocation modules in the HGformer framework. From top to bottom, it consists of the Enhanced Graph Transformer Encoder (EGTE), the Planner Agent decoding stage using a pointer network with multi-head cross-attention, and the Transfer Agent decoding stage using dual-scale mixed encoding.}
  \label{fig 2}
\end{figure*}

The following sections detail the architecture of EGTE, the design of the Planner and Transfer Agent decoders, and the joint training procedure (see Fig.~\ref{fig 2}).

\begin{algorithm}[ht]
\caption{Inference Process of HGformer}
\label{alg1}
\SetKwInOut{Input}{Input}
\SetKwInOut{Output}{Output}
\Input{Planner(), Transfer(), one case $\mathcal{G}$}
\Output{$U_r^*, U_b^*$ and the occupation of each node on the graph}
$S_{0}^{b} \leftarrow \mathrm{ReluPlanner}(\mathcal{G})$\;
$S_{0}^{r} \leftarrow \mathrm{Planner}(S_{0}^{b},\,\mathcal{G}\mid \theta^{P})$\;
Update the resource status of each node\;
\While{termination conditions not met}{
  $S_{t}^{b} \leftarrow \mathrm{ReluTransfer}\bigl(\mathcal{G},\,S_{t-1}^{b},\,S_{t-1}^{r}\bigr)$\;
  $S_{t}^{r} \leftarrow \mathrm{Transfer}\bigl(\mathcal{G},\,S_{t}^{b},\,S_{t-1}^{r}\mid \theta^{T}\bigr)$\;
  Update the resource status of each node\;
  $t \leftarrow t + 1$\;
}
\end{algorithm}

\subsection{Enhanced Graph Transformer Encoder (EGTE)}\label{subsection3.1}
EGTE combines the global modeling power of self-attention with a shortest path distance (SPD) based structural bias to effectively capture long-range dependencies between nodes and the dynamic resource distribution features.

\subsubsection{Node state feature representation} We design a feature vector that captures the key state of each node in the context of the game’s dynamics. Considering the “winner-keeps, loser-loses” resource rule: after each local confrontation at a node, the side with fewer resources loses them while the side with more retains theirs. Based on this adversarial dynamic (and to simplify learning), we represent each node’s current state as a three-dimensional feature vector:
\begin{equation}
    x_i(t)=[v_i,\mathscr{s}_i^r(t),\mathscr{s}_i^b(t)]
\end{equation}
where $v_i$ is the node’s value, and $\mathscr{s}_i^r(t)$ and $\mathscr{s}_i^b(t)$ are the amounts of Red and Blue resources currently at node $i$ (if either side has lost control of the node in a previous round, its corresponding resource amount is set to 0). This triplet captures the key state information of node $i$ in the game. To enhance the discriminative power of the features, we also include the node’s degree $deg_{i}=|\mathcal{N}_{i}|$ as a measure of importance. We then employ two sets of learnable mappings to lift these features to the model’s hidden dimension: one multi-layer perceptron (MLP) $\mathbf{MLP_1}\in\mathbb{R}^{M\times d}$ maps the original 3-d feature $x_i(t)$ into the hidden dimension, and another embedding $\mathbf{Emb}_{1}\in\mathbb{R}^{1\times d}$ maps the scalar degree into the same dimension; the two are added together. Denote the resulting initial hidden representation of node $i$ as $h_i^0$. Formally:
\begin{equation}
    h_i^0=\mathbf{MLP_1}\left(x_i(t)\right)+\mathbf{Emb_1}(deg_i)
\end{equation}

 Through this processing, each node obtains an initial feature vector at the encoder input that fuses its degree centrality and current resource state. 

\subsubsection{SPD Structural Bias}In graph-structured data, nodes do not possess a natural linear order, making it difficult for standard Transformer architectures to capture structural distances and topological relationships. To address this limitation, we introduce a shortest path distance (SPD)-based structural bias into the attention mechanism. By embedding SPD information into the attention score calculation, the model is guided to focus more on node pairs that are structurally proximate, thereby enhancing its ability to model local interactions and effectively encode graph topology.

Specifically, we precompute the shortest path length $\delta(i,j)$ between every pair of nodes. If two nodes are not connected, we set the distance as $\delta(i,j)=\infty$. We then introduce a learnable bias $\mathscr{b}_{\delta(i,j)}\in\mathbb{R}$ for each distance value. In the self-attention calculation, for each node pair $(i,j)$
, in addition to using the conventional "query-key" similarity, we add the corresponding SPD bias $\mathscr{b}_{\delta(i,j)}$.

\subsubsection{Multi-Head Structural Bias Self-Attention (SPD-MHA)} The EGTE consists of $L$ layers of Transformer encoders, with each layer comprising multi-head self-attention (MHA) and feed-forward network (FFN) sub-layers, augmented with layer normalization and residual connections to ensure stable training. For the $l$-th layer, the process is as follows: 

Let $h_{i}^{(l-1)}$ denote the output of the $l$-th layer for node iii. We first compute the query, key, and value vectors for the $l$-th layer: 
\begin{equation}
    \begin{aligned} 
          q_{i}^{(l,h)}=h_{i}^{(l-1)}W_{Q}^{E(l,h)},\\
          k_{j}^{(l,h)}=h_{j}^{(l-1)}W_{K}^{E(l,h)},\\
          v_{j}^{(l,h)}=h_{j}^{(l-1)}W_{V}^{E(l,h)},
      \end{aligned} 
\end{equation}
where $H$ is the number of attention heads, and $W_Q^{E(l,h)},\cdot W_K^{E(l,h)},W_V^{E(l,h)}\in\mathbb{R}^{d\times(\frac{d}{H})}$ are the projection matrices. For each pair of nodes $(i,j)$ and each head $h$, the unnormalized attention score with the SPD bias is computed as: 
\begin{equation}
    e_{ij}^{(l,h)}=\frac{\langle q_{i}^{(l,h)},k_{j}^{(l,h)}\rangle}{\sqrt{d/H}}+\mathscr {b}_{\delta(i,j)}.
\end{equation}
where $\langle\cdot,\cdot\rangle$denotes the inner product (or dot product) between two vectors.

The unnormalized scores of each attention head are passed through the SoftMax function to obtain the attention weights $\alpha_{ij}^{(l,h)}$; these weights are then used to compute a weighted sum of the value vectors, resulting in the output representation of each attention head. 
\begin{equation}
h_i^{\prime{(l,h)}}=\sum_{j\in\mathcal{N}}\alpha_{ij}^{(l,h)}v_j^{(l,h)}.
\end{equation}

The outputs of the $H$ attention heads are concatenated and projected back to a $d$-dimensional space, forming the output of the multi-head attention (MHA) sublayer. A residual connection is then added, followed by layer normalization (LayerNorm): by incorporating the residual connection and layer normalization, the final expression is given as follows. 
\begin{equation}
    h_i^{\prime{(l)}}=\mathrm{LN}(h_i^{(l-1)}+\mathrm{MHA}(h_i^{(l-1)}))
\end{equation}
The next step involves passing the result through two layers of feed-forward networks (FFN) with residual connections and layer normalization: 
\begin{equation}
h_{i}^{(l)}=\mathrm{LN}\left(h_{i}^{\prime(l)}+W_{2}^{E(l)}\sigma\left(W_{1}^{E(l)}h_{i}^{\prime(l)}\right)\right)
\end{equation}
where  $W_{1}^{E(l)}\in\mathbb{R}^{d\times d_{ff}}$,  $W_2^{E(l)}\in\mathbb{R}^{d_{ff}\times d}$, $d_{ff}$ denotes the dimensionality of the feed-forward network. 

This mechanism adjusts attention weights not only based on feature similarity but also on the topological proximity of nodes. As a result, nodes that are topologically closer are assigned relatively higher attention weights, enhancing the local interaction modeling.

\subsubsection{Virtual Node Aggregation of Global Information} During the process of node feature embedding, we adopt the idea from BERT’s [CLS]\cite{35} token to aggregate global information. We introduce an additional virtual node $\mathscr{v}_{0}$, which is considered adjacent to all real nodes in the graph (i.e., $\delta(v_{0},j)=1,\forall j\in\mathcal{N}$). For every node $i$, we assign a learnable special bias $h_{v_{0}}\in\mathbb{R}^{1\times d}$ for the virtual node.

In the Transformer encoding process, $\mathscr{v}_{0}$ participates in the multi-head self-attention calculations, effectively acting as a global "aggregator." It collects information from all nodes at each layer and updates its representation $h_{v_{0}}^{(l)}$. After $L$ layers, $h_{v_{0}}^{(L)}$ can be treated as the global embedding of the entire graph. Meanwhile, each real node’s representation $h_{i}^{(L)}$ combines both its local features and the global adversarial information propagated through multi-hop attention.

The final output $\left\{h_{i}^{(L)}\right\}_{i\in\mathcal{N}}$ of the EGTE encoder is used as input features for the subsequent strategy decoding module, with the global embedding $h_{v_{0}}^{(L)}$ providing a summary expression of the entire resource node state. 

The EGTE algorithm is outlined in Algorithm \ref{alg2}, Lines 1–3 map the original node features and degree information through MLPs and embedding layers, introducing the virtual node. Line 4 precomputes the shortest path distances between node pairs and assigns a learnable bias for each distance. Lines 5–11 apply the SPD-biased multi-head self-attention and feed-forward network sub-layers across all nodes in the stacked Transformer layers.

\begin{algorithm}[!htbp]
\caption{Enhanced Graph Transformer Encoder}
\label{alg2}
\SetKwInOut{Input}{Input}
\SetKwInOut{Output}{Output}

\Input{$\mathcal{G} = (\mathcal{V},\mathcal{E},\mathcal{W})$, 
 $\{x_i(t)\}_{i\in\mathcal{V}}$, $\{\,\deg(i)\}_{i\in\mathcal{V}}$, 
 $\delta(\cdot,\cdot)$, $L$, $d$.}
\Output{$\{h_i^L\}_{i\in\mathcal{V}}$, $h_\mathscr {v}^L$.}

\BlankLine
\textbf{// Step A: Node feature initialization}\;
Add a virtual node $\mathscr{v}_0$ to $\mathcal{N}$ and set $\delta(\mathscr{v}_0,i)=1,\;\forall i\in\mathcal N$\;

Compute for each node $i\in\mathcal N$:
$h_i^0 \leftarrow \mathrm{MLP}_1(x_i(t)\bigr)\;+\;\mathrm{Emb}_1(deg(i)\bigr)$\;

Initialize the virtual node: $h_{\mathscr{v}_0} \leftarrow \mathbf{0}_d$\;

\BlankLine
\textbf{// Step B: SPD-based structure bias}\;
\ForEach{pair $(i,j)\in (\mathcal{N}\cup\{\mathscr{v}_0\})\times(\mathcal{N}\cup\{\mathscr{v}_0\})$}{
  Compute structural bias $\mathscr{b}_{\delta(i,j)}$\;
}

\BlankLine
\textbf{// Step C: Layer-by-layer Transformer encoding}\;
\For{$l \leftarrow 1$ \KwTo $L$}{
  \ForEach{head $h$ and node $i$}{
    $q_i^{(l,h)} \leftarrow W_Q^{(l,h)}\,h_i^{l-1}$\;
    $k_j^{(l,h)} \leftarrow W_K^{(l,h)}\,h_j^{l-1}$\quad for all $j$\;
    $v_j^{(l,h)} \leftarrow W_V^{(l,h)}\,h_j^{l-1}$\;
    Compute attention scores:
      $e_{ij}^{(l,h)} \leftarrow \frac{\langle q_i^{(l,h)},k_j^{(l,h)}\rangle}{\sqrt{d/H}}
      + \mathscr{b}_{\delta(i,j)}$
    \;
    $\alpha_{ij}^{(l,h)} \leftarrow \mathrm{Softmax}_j\bigl(e_{ij}^{(l,h)}\bigr)$\;
  }
  Compute $\mathrm{MHA}$ output and update:
    $ h_i'^{\,(l)} \leftarrow \mathrm{LN}\bigl(h_i^{(l-1)} + \sum_h \sum_j \alpha_{ij}^{(l,h)} v_j^{(l,h)}\bigr)$
  \;
    $ h_i^{(l)} \leftarrow \mathrm{LN}\bigl(h_i'^{\,(l)} + \mathrm{FFN}(h_i'^{\,(l)})\bigr)$
  \;
}
\end{algorithm}

\subsection{Planner Agent Decoder: Sequential Resource Allocation Strategy}\label{subsection3.2}

The initial allocation is inherently sequential: Red must decide where to invest and how much to commit at each step.  
A naive “one–shot’’ decoder that outputs the full allocation vector in a single pass ignores decision inter-dependencies and therefore generalizes poorly.  

We instead adopt a hybrid Pointer–Attention decoder.  
Following the inductive bias of Pointer Networks \cite{36}, the decoder iteratively selects one target node at a time, while multi-head cross-attention \cite{34} injects global context to guide each choice.  
After every selection the remaining budget is updated, and the node embeddings are refreshed, enabling subsequent decisions to adapt to earlier commitments.  
This step-wise procedure preserves dependency structure and yields markedly stronger generalization than static, one-shot allocation.

Concretely, the EGTE encoder outputs a global graph embedding \( h_{\mathscr{v}_0}^L \)) and node-level embeddings. The Planner decoder leverages these to perform up to \( N \) decoding steps (where the process may terminate early if resources are exhausted or no viable actions remain). Let \( \mathscr{r}_\mathscr{n}^r \) denote Red’s remaining resources at step \( \mathscr{n} \)-th decoding step (\( \mathscr{n} = 1, 2, \ldots, N \)), initialized as \( \mathscr{r}_1^r = \mathcal{R}^r \).

\textbf{Step \( \mathscr{n} = 1 \)}:  
Before any node is selected, there is no "previous node." We introduce a fixed placeholder embedding \( h_{\text{prev}} \) to represent the "previous node" at the first step. By combining this placeholder embedding with the current remaining resource \( \mathscr{r}_1^r \), we define a local context projection function \( W_{\text{step}} \in \mathbb{R}^{(d+1) \times d} \) via a feed-forward network to generate the local context vector for the first decoding step:
\begin{equation}
    c_{\text{local}}^{(1)} = W_{\text{step}}([h_{\text{prev}}, \mathscr{r}_1^r]),
\end{equation}
where \([\,\cdot\,,\,\cdot\,]\) denotes the horizontal concatenation. Meanwhile, the global embedding \( h_{\mathscr{v}_0}^{(L)} \) is projected through a linear transformation \( W_{\text{fix}} \in \mathbb{R}^{d \times d} \) to obtain the global context vector:
\begin{equation}
    c_{\text{global}} = W_{\text{fix}} h_{\mathscr{v}_0}^{(L)}.
\end{equation}

We then sum the global and local context vectors to form the combined context for step 1:
\begin{equation}
    c^{(1)} = c_{\text{local}}^{(1)} + c_{\text{global}}.
\end{equation}

\textbf{Step \( \mathscr{n} \) (\( \mathscr{n} \geq 2 \))}:  
Suppose that at step \( \mathscr{n} - 1 \), the Planner selected node \( j \). Then, at step \( \mathscr{n} \), we use the embedding \( h_j^{(L)} \) of node \( j \) along with the current remaining resource \( \mathscr{r}_\mathscr{n}^r \) as inputs to the local context function:
\begin{equation}
    c_{\text{local}}^{(\mathscr{n})} = W_{\text{step}}([h_j^{(L)}, \mathscr{r}_\mathscr{n}^r]).
\end{equation}

This local context is then combined with the unchanged global context to produce the final context vector for this step:
\begin{equation}
c^{(\mathscr{n})} = c_{\text{local}}^{(\mathscr{n})} + c_{\text{global}}.
\end{equation}

In this way, each decoding step considers the global resource node context while dynamically incorporating the most recent resource allocation history.

Next, the Planner decoder treats the combined context vector $c^{(n)}$ at each decoding step as a special query vector. Using this query, we perform multi-head cross-attention over all node embeddings $\{ h_i^{(L)} \}_{i \in \mathcal{N}}$ to compute the suitability scores for selecting each node as the next target. Specifically, parameterized matrices $W_q^P$, $W_k^P$, and $W_v^P \in \mathbb{R}^{d \times d}$ are used to construct the query, key, and value vectors as follows.
\begin{equation}
q_c^{(\mathscr{n})} = W_q^P c^{(\mathscr{n})}, \quad k_i = W_k^P h_i^{(L)}, \quad v_i = W_v^P h_i^{(L)}.
\end{equation}

The attention score for node $i$ at step $\mathscr{n}$ is computed as
\begin{equation}
e_i^{(\mathscr{n})} = \frac{\langle q_c^{(\mathscr{n})}, k_i \rangle}{\sqrt{d_k}}.
\end{equation}

Let $\mathcal{N}^{(\mathscr{n})}$ denote the set of nodes already selected before step $n$. For any $i \in \mathcal{N}^{(\mathscr{n})}$, a masking mechanism sets $e_i^{(\mathscr{n})} = -\infty$ to prevent repeated selection. For the remaining nodes, the scores are passed through a softmax function to yield the selection probability for node $i$ at step $\mathscr{n}$:
\begin{equation}
p_i^{(\mathscr{n})} = \frac{\exp(e_i^{(\mathscr{n})})}{\sum_{j \in \mathcal{N} \setminus \mathcal{N}^{(\mathscr{n})}} \exp(e_j^{(\mathscr{n})})}.
\end{equation}

This probability distribution guides the Planner to select the next node either greedily or via sampling.

Once node $j$ is selected at step $\mathscr{n}$, the red agent allocates resources to it and updates the remaining resource as:
\begin{equation}
r_\mathscr{n}^r = r_{\mathscr{n}-1}^r - s_j^b(0),
\end{equation}
where $s_j^b(0)$ denotes the resource invested in node $j$, typically set to the minimum required to overcome Blue’s defense at that node. The embedding $h_j^{(L)}$ of the selected node is then used as the “previous node” embedding for computing $c_{local}^{(\mathscr{n}+1)}$ in the next step. This iterative process continues until the Red agent exhausts its available resources or the predefined step limit is reached.

For clarity in subsequent descriptions, we treat the above Planner Agent as a policy network with parameters $\theta^P$. Specifically, $\theta^P$ consists of two submodules: the encoder parameters $\theta_E^P$ based on EGTE, and the sequential decoder parameters $\theta_d^P$:

Unlike one–shot schemes, our Planner decoder treats initialization as a sequential planning task, faithfully reflecting budget limits and inter–node dependencies. A learned query vector—obtained by fusing EGTE’s global embedding with step-wise decision history—injects both macro context and local dynamics into every decoding step. Consequently the decoder prioritizes nodes and allocates resources more accurately, yielding higher strategic gains and stronger generalization in multi-stage scenarios.

\subsection{Transfer Agent Decoder: Local-Global Hybrid Resource Transfer Strategy}\label{section3.3}
During dynamic transfers, Red must react in milliseconds to shifting neighbourhood states while still reasoning over the global graph. Resource flows are highly local—nodes continually reinforce nearby positions or mount targeted moves—yet long-horizon success hinges on synchronising these micro-actions with network-wide objectives. EGTE supplies rich long-range context, but its global pooling attenuates the tightly coupled signals that drive fine-grained moves. We therefore introduce a dual-scale encoder that augments EGTE with a lightweight, edge-conditioned GAT module, fusing panoramic awareness with high-resolution local dynamics and enabling precise, globally coherent reallocations.

\subsubsection{Global-Local Dual-Scale Encoding}
Specifically, during the encoding phase of the Transfer Agent, we first adopt the EGTE encoder to represent the current global network state, obtaining global embeddings for all nodes $\{ h_i^{(L)} \}_{i \in \mathcal{N}} \in \mathbb{R}^d$ and the entire graph's global embedding $h_{v_0}^{(L)} \in \mathbb{R}^d$. While EGTE provides features enriched with long-range dependencies, we additionally employ a fine-grained multi-head Graph Attention Network v2 (GATv2) module \cite{37} to specifically capture dynamic features in each node's local neighborhood.

The GATv2 module first applies a learnable linear transformation $W_G^T \in \mathbb{R}^{d \times d}$ to each node's feature $x_i(t)$ to project it into a unified feature space:

\begin{equation}
z_i = W_G^T x_i(t)
\end{equation}

Then, for each neighboring pair $i$-$j$ (where $j \in \mathcal{N}_i$), we define the attention score:

\begin{equation}
e_{ij} = \text{LeakyReLU}(W_l^T z_i + W_r^T z_j),
\end{equation}
where $\text{LeakyReLU}$ denotes the leaky rectified linear activation function, and $W_l^T, W_r^T \in \mathbb{R}^{d \times d}$ are trainable matrices used to extract the attention contributions of nodes $i$ and $j$, respectively. 

To restrict attention to legitimate neighbours, we set $e_{ij}=-\infty$ whenever $j\notin\mathcal N_i$, so that the subsequent softmax normalisation is confined to $\mathcal N_i\cup\{i\}$.  The resulting local embedding is
\begin{equation}
h_i^{\text{loc}} = \sigma\left(\sum_{j \in \mathcal{N}(i) \cup \{i\}} \text{Softmax}(\alpha_{ij}) z_j \right),
\end{equation}
where $\sigma(\cdot)$ denotes a point-wise non-linearity. 

Relying solely on local aggregation is still insufficient to fully capture the trends of global resource flow. To address this limitation, we introduce a learnable vector $h_{v_a} \in \mathbb{R}^d$ on top of the multi-layer GAT outputs to extract global information. Specifically, $h_{v_a}$ is treated as a query vector. Each node $i$ computes a similarity score between its local representation $h_i^{\text{loc}}$ and $h_{v_a}$, which is then used for weighted aggregation to produce a global view from the GAT perspective:
\begin{equation}
h_{GAT}^{glob} = \sum_{i=1}^N \text{Softmax}\left(\langle h_i^{loc},h_{v_a} \rangle\right) h_i^{loc}.
\end{equation}

Next, we fuse the node embeddings from the global perspective (i.e., from the EGTE encoder) and the local perspective (i.e., from the GAT). For each node $i$, its final representation $h_i^*$ is computed by summing its global embedding $h_i^{(L)}$ and local embedding $h_i^{\text{loc}}$:
\begin{equation}
h_i^* = h_i^{(L)} + h_i^{loc}.
\end{equation}

Meanwhile, the final graph-level representation is obtained by concatenating the EGTE-derived global embedding $h_{v_0}^{(L)}$ and the GAT-derived global embedding $h_{GAT}^{glob}$, followed by a linear transformation using $W_A^T \in \mathbb{R}^{2d \times d}$:
\begin{equation}
h_{glob}^* = W_A^T [h_{v_0}^L, h_{GAT}^{glob}].
\end{equation}

Through this global-local hybrid encoding process, we construct a state representation for the Transfer phase that integrates both long-range dependencies and local dynamic features.

\subsubsection{Resource Transfer Decoding}  
With the above state representations in place, the Transfer Agent employs a multi-layer Transformer decoder to generate concrete resource transfer decisions for each node. The objective is: for each red node $i$ that still has remaining resources (i.e., $\mathscr{s}_i^r(t) > 0$), determine which neighboring nodes to transfer resources to, and in what proportion. We treat each source node $i$'s decision as a local attention-based allocation problem: node $i$ serves as the query, while its neighbors in $\mathcal{N}_{(i)}$ serve as candidates. Attention scores from $i$ to each neighbor $j \in \mathcal{N}_{(i)}$ determine the proportional distribution of its outgoing resources.

Specifically, for any source node $i$, we first construct its query vector:
\begin{equation}
q_i^{trans} = W_q^T [h_i^*, h_{glob}^*],
\end{equation}
where $W_q^T \in \mathbb{R}^{2d \times d}$ is the projection matrix that fuses the node's final embedding with the global graph embedding. For each neighbor $j \in \mathcal{N}(i)$, its final embedding $h_j^*$ is used as the key vector, which is projected using another matrix $W_k^T$:
\begin{equation}
k_j^{trans} = W_k^T h_j^*.
\end{equation}

The attention-based transfer score from node $i$ to neighbor $j$ is computed as:
\begin{equation}
e_{i \rightarrow j} =
\begin{cases}
\frac{\langle q_i^{trans} , k_j^{trans}\rangle}{\sqrt{d}}, & \text{if } j \in \mathcal{N}(i), \\
-\infty, & \text{otherwise}.
\end{cases}
\end{equation}

At each decision step the Transfer Agent converts the attention scores into a probability vector
$\mu_{i\rightarrow j}^{r}$ by a softmax, yielding a proportionate transfer policy from node~$i$ to its neighbours.  The actual flow is then
$\Delta\mathscr{s}_{i\rightarrow j}^{r}(t)=\mu_{i\rightarrow j}^{r}(t)\,\mathscr{s}_{i}^{r}(t),$
so that node~$i$ dispatches the fraction prescribed by $\mu_{i\rightarrow j}^{r}$ while retaining the remainder.  

Once every source node has determined its transfers, Red’s post-move resources are updated according to~\eqref{trans}.  Blue simultaneously executes its own movement rule, producing the next-round state $\{\mathscr{s}_{i}^{r}(t+1)\}_{i\in\mathcal N}$.  The game then proceeds to the subsequent confrontation round.

Similarly, we model the Transfer Agent as a policy network with parameters denoted by $\theta^T$. 
\begin{equation}
\text{Transfer:} \quad \theta^T = \{\theta_E^T, \theta_G^T, \theta_d^T\}.
\end{equation}

The Transfer Agent is designed to integrate both local and global features. At each decision step, nodes determine outward resource transfers based on their own state and global context, aligning local tactical actions (e.g., neighborhood reinforcements) with global strategic goals (e.g., maximizing overall node control advantages). 

\subsection{Layered Feedback Reinforced Training}

Layered Feedback Reinforced Training (LFRT) jointly optimizes the high-level Planner and the low-level Transfer policies in three successive phases.

\subsubsection{Phase A: REINFORCE Pre-training of the Planner}
With the Transfer layer disabled, the Planner policy $\pi_{\theta^{P}}$ is optimized by REINFORCE \cite{38}.  
Given the initial Blue allocation $\mathcal{S}_0^{b}$ and graph $\mathcal G$, the Planner outputs a complete allocation sequence $a_0$.  
Its return is
\[
R_0^P=\sum_{i=1}^{N}\!\mathbb{I}\!\bigl(\mathscr{s}_i^r\!\ge\!\mathscr{s}_i^b\bigr)v_i,
\]
and the gradient
\[
\nabla_{\theta^{P}}\!J= \mathbb{E}\bigl[R_0^P\nabla_{\theta^{P}}\!\log\pi_{\theta^{P}}(a_0)\bigr],
\]
estimated with generalized advantage estimation (GAE) and a learned value baseline.  
After convergence, the Planner alone already yields competitive node control.

\subsubsection{Phase B: PPO Pre-training of the Transfer Layer}
Fixing the Planner, the Transfer policy $\pi_{\theta^{T}}$ is trained with PPO \cite{39}.  
The MDP state at step $t$ is $s_t=(h_{\mathrm{glob}}^{*}(t),\{h_i^{*}(t)\}_{i=1}^{N})$, and an action $a_t$ is a transfer matrix $\mu_{i\rightarrow j}^r(t)$.  
The composite reward
\[
R_t^{T}= \Delta C_t+\alpha_s\frac{R_t^{r}-R_t^{b}}{R_t^{r}+R_t^{b}}
        -\frac{\sum_{i}c^{ij}(t)}{\sum_{i}\mathscr{s}_i^{r}(t)},
\]
where $\Delta C_t$ is the change in control score and $\alpha_s\!=\!0.1$, guides resource-aware manoeuvres.  
PPO optimizes the clipped surrogate loss, value loss, and entropy regularizer across parallel simulations.

\subsubsection{Phase C: Hierarchical Feedback Fine-tuning}
Let $U_{r}(\hat{\theta}^{T})$ be the average cumulative reward achieved by the converged Transfer policy.  
The Planner is refined by maximizing
\[
J'(\theta^{P})=\mathbb{E}\!\bigl[R_0^{P}+\lambda U_{r}(\hat{\theta}^{T})\bigr],
\]
where $\lambda$ balances static and dynamic objectives.  
A few REINFORCE updates with this augmented reward align high-level allocations with low-level capabilities, closing the coordination loop.

Algorithm \ref{alg3} summarizes LFRT: Phase A (lines 2–9), Phase B (10–28), and Phase C (31–38) are executed sequentially, yielding a pair of mutually adapted hierarchical policies.
\begin{algorithm}[htbp]
\caption{Layered Feedback Reinforced Training (LFRT)}
\label{alg3}
\SetKwInOut{Input}{Input}
\SetKwInOut{Output}{Output}

\Input{Initial Planner policy parameters $\theta_P$, initial Transfer policy parameters $\theta_T$}
\Output{Optimized Planner parameters $\hat\theta_P$ and Transfer parameters $\hat\theta_T$}
\BlankLine
\textbf{Phase A: Pre-train Planner (REINFORCE)}\;
\For{each iteration}{
  Sample a mini-batch $\{(S_0^{b,i}, G^i)\}_{i=1}^M$\;
  Generate initial actions $\{a_0^i\}_{i=1}^M \sim \pi_{\theta_P}$\;
  Compute returns $\{R_0^{P,i}\}_{i=1}^M$\;
  Compute policy gradient: 
    $\nabla_{\theta_P}J
    = \frac{1}{M} \sum_{i=1}^M R_0^{P,i}\,\nabla_{\theta_P}\log \pi_{\theta_P}(a_0^i \mid \mathcal{S}_0^{b,i}, \mathcal{G}^i)$\;
  Update Planner policy:$\theta_P \gets \theta_P + \alpha\,\nabla_{\theta_P}J$\;

  Store pretrained Planner as $\tilde\theta_P \gets \theta_P$\;
}
\BlankLine
\textbf{Phase B: Pre-train Transfer (PPO)}\;
\For{epoch = 1 \KwTo $N_1$}{
  Sample $\{(\mathcal{S}_0^{b,i},\mathcal{G}^i)\}_{i=1}^M$\;
  Use $\tilde\theta_P$ to generate $\{a_0^i\}$ and form  $\{s_1^i\}$\;
  \For{$t=1$ \KwTo $\mathcal{T}$}{
    \For{$i=1$ \KwTo $M$}{
      Sample $a_t^i\sim \pi_{\theta_T}(\cdot\mid s_t^i)$, observe reward $R_t^{T,i}$ and next state $s_{t+1}^i$\;
      Store $(s_t^i, a_t^i, R_t^{T,i})$ in trajectory $\tau^i$\;
    }
  }
  Compute advantages $\{\hat A_t^i\}$ for each trajectory\;
  \For{each $(i,t)$}{$\rho_t^i \gets {\pi_{\theta_T}(a_t^i\!\mid\!s_t^i)}/{\pi_{\theta_T^\text{old}}(a_t^i\!\mid\!s_t^i)}$\;}
  Compute clipped surrogate loss:
    $L^\text{CLIP}
    = \frac{1}{M}\sum_{i=1}^M\sum_{t=1}^T
      \min\bigl(\rho_t^i\,\hat A_t^i,\;\mathrm{clip}(\rho_t^i,1-\epsilon,1+\epsilon)\,\hat A_t^i\bigr)$\;
  Compute entropy loss $L^S$ and value loss $L^V$:
  $L^S = -\frac{1}{M}\sum_{i,t}\sum_a \pi_{\theta_T}(a\!\mid\!s_t^i)\,\log\pi_{\theta_T}(a\!\mid\!s_t^i),
    \quad$
    $L^V = \frac{1}{M}\sum_{i,t}\bigl(V(s_t^i) - G_t^i\bigr)^2$\;
  Total loss:
    $L^\text{total} = -L^\text{CLIP} + c_1\,L^V - c_2\,L^S$\;
  Update Transfer policy:$\theta_T \gets \theta_T - \beta\,\nabla_{\theta_T}L^\text{total}$\;
}
Store pretrained Transfer as $\hat\theta_T \gets \theta_T$\;
\BlankLine
\textbf{Phase C: Cross-layer Feedback to Planner}\;
Sample $\{(\mathcal{S}_0^{b,i},\mathcal{G}^i)\}_{i=1}^M$\;
Simulate full trajectories with $(\tilde\theta_P, \hat\theta_T)$, record dynamic returns $U_r^i$\;
\For{each sample $i$}{
  Compute enhanced return $\tilde R^i = R_0^{P,i} + \lambda\,U_r^i$\;
}
Compute feedback gradient:
  $\nabla_{\tilde\theta_P}J
  = \frac{1}{M}\sum_{i=1}^M \tilde R^i
    \,\nabla_{\tilde\theta_P}\log \pi_{\tilde\theta_P}(a_0^i \mid S_0^{b,i},G^i)$\;
Update Planner policy:$\tilde\theta_P \gets \tilde\theta_P + \alpha\,\nabla_{\tilde\theta_P}J$\;
\Return{$(\hat\theta_P, \hat\theta_T)$}
\end{algorithm}

This layered training procedure enables effective transmission of optimization signals from the lower-level policy to the upper-level strategy, ensuring close coordination between Red’s initial deployment and subsequent dynamic scheduling. Experimental results demonstrate that incorporating LFRT yields substantially higher global returns than independently training the two layers, thereby validating the effectiveness of the proposed collaborative training mechanism.

\section{Experiments}\label{section4}
We assess HGformer on the graph-constrained two-stage Colonel Blotto game through a comprehensive experimental suite.  
Because no end-to-end baseline exists for this setting, we construct comparators for each stage independently—static pre-allocation and dynamic transfer—as well as their combinations over the full task.  
Ablation studies further isolate the contribution of every key module.  All methods are trained and tested under identical environments and hyper-parameters to ensure a fair and reproducible comparison.

\subsection{Experimental Setup}\label{section4.1}
\subsubsection{Graph data generation and environment}  We generate random graph scenarios with a graph size $N$ ranging from 10 to 70. Each node’s value $v_i$ is sampled uniformly from [0,1]. Blue’s initial total resource $\mathcal{R}^{b}$ is set to $5N$, while the Red receives half that amount: $\mathcal{R}^{r} = 0.5\mathcal{R}^{b}$simulating an asymmetric scenario. All algorithms are evaluated on the same set of random seeds to ensure consistent and fair comparisons.

\subsubsection{Model training process and parameters} HGformer is trained in three stages:  Planner Agent (500 iterations, evaluation every 1 iteration), Transfer Agent (3000 iterations, with 10 PPO epochs per iteration, evaluation every 10 iterations), joint optimization with feedback (LFRT, 2000 iterations, evaluation every 10 iterations). Both the Planner and Transfer modules use the EGTE with $L=3$  layers and $H=4$ attention heads. The node embedding dimension is 32, and the FFN dimension is 64. Learning rates are: Planner $10^{-3}$, Transfer policy/value networks $5\times10^{-4}$. PPO-specific parameters:clipping coefficient = 0.2, entropy regularization = 0.01, value loss coefficient = 0.5, mini-batch size = 10 episodes. 
Blue’s policy is rule-based: initial allocation is proportional to node values; dynamic reinforcement prioritizes recaptured high-value neighbors. In LFRT, the Transfer Agent’s cumulative utility is incorporated as feedback to the Planner, with a feedback weight $\lambda=0.5$

\subsubsection{Performance metrics} We adopt three metrics for performance evaluation: (1) Red utility $\overline{U}r$: total value of nodes held by Red at the end (higher is better); (2) Transfer cost $\bar{C}^{r}$: cumulative transfer expenditure (lower is better); and (3) Decision latency $\bar{E}^{r}$: time taken to compute a complete strategy (lower indicates higher deployability).

This unified setup ensures fair, consistent evaluation across all components and algorithms.

\subsection{Initial Allocation Stage Experiments} \label{section4.2}
This study isolates the Planner Agent, treating the pre-allocation task as a combinatorial optimisation problem under graph and budget constraints.  The objective is to maximise Red’s captured value given Blue’s known initial layout.

To comprehensively assess the Planner Agent’s performance, we have designed three baseline algorithms and three ablation model variants for benchmarking.   Greedy heuristic—ranks nodes by value density (node value divided by Blue’s initial resource) and allocates Red’s resources greedily based on this ratio until depletion.  Simulated Annealing  \cite{40}—a meta-heuristic search method that iteratively perturbs the allocation vector, accepting changes probabilistically to explore high-quality solutions.  Mixed Integer Linear Programming (MILP)—formulates the task as an integer linear program and solves it using a commercial solver to obtain a near-optimal allocation. 

The ablation models are variants of the Planner with specific components removed or altered to isolate their contributions:  GNN-P: replaces the EGTE encoder with a standard Graph Convolutional Network (GCN), while retaining the decoder. GTE-P: removes SPD structural bias and node degree features, reverting to a vanilla Transformer encoder to test the EGTE's structural enhancements. GAE-P: uses EGTE for encoding but replaces the sequential decoder with a one-shot proportional output, assessing the importance of sequential decision-making. All models are trained and evaluated under identical conditions.  One hundred random graphs constitute the test suite.  For each graph Blue’s allocation is fixed, and each method computes Red’s deployment, from which base payoff and decision time are measured.  Reported results are averages over all instances.

\begin{figure*}[h]
  \centering
  \includegraphics[width=0.95\textwidth]{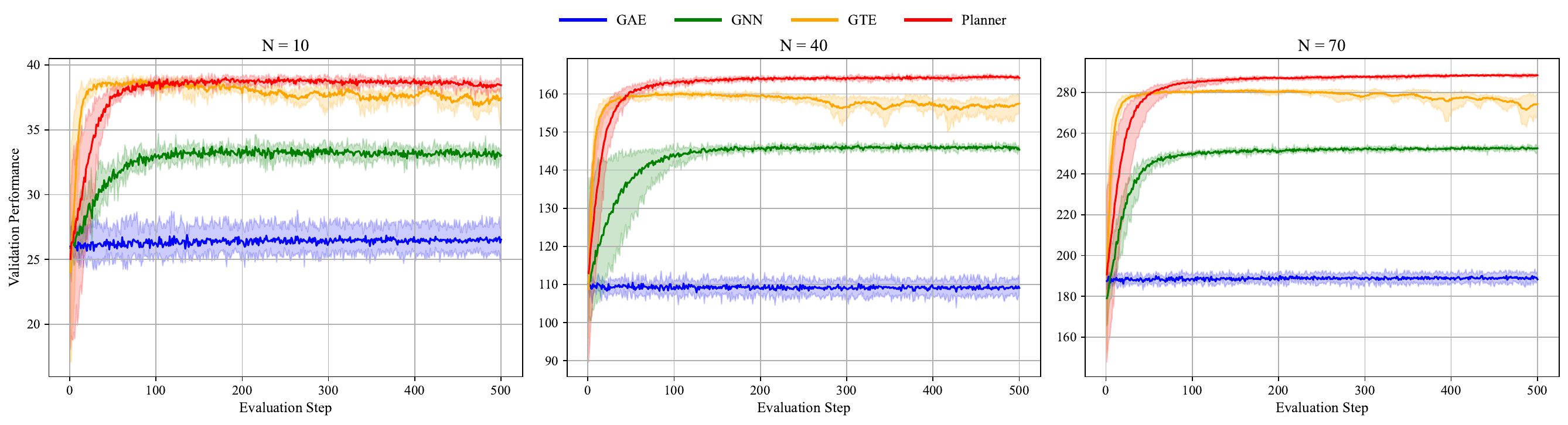}
  \caption{Training convergence curves for Planner ablation experiments: (a) $N=10$; (b) $N=30$; (c) $N=70$. The full HGformer Planner (red solid line) converges the fastest and consistently achieves the highest validation reward across all graph sizes. Replacing the enhanced Transformer encoder with a vanilla GNN (green line) or with a GAE baseline (blue line) leads to markedly lower rewards. A single-shot variant that removes the sequential planning bias (orange line, “GTE”) reaches only the second-best reward and converges more slowly than the full Planner. Shaded areas show the mean ± 1 s.e.m. over 5 runs.}
  \label{curves}
\end{figure*}

Table \ref{planner1}  show that the Planner delivers consistently higher utility than heuristics and SA, and matches MILP on small graphs while remaining orders-of-magnitude faster.  On $N=70$ graphs, MILP requires tens of seconds, whereas the Planner responds in milliseconds, making it practical for real-time use.  
Ablation curves in Fig.\ref{curves} confirm that both EGTE’s structural bias and the sequential pointer decoder are indispensable: replacing either component reduces final utility and slows convergence.  The full Planner thus achieves the best trade-off between accuracy and scalability.

\begin{table*}[htbp]
  \centering
  \caption{Performance of different algorithms in the initial allocation stage (Red’s average utility and decision time) across various graph sizes.}
  \label{planner1}
  \begin{tabular}{ccrrrrrrr}
    \toprule
    Size & Metric & Greedy       & SA            & MILP          & GNN-P       & GTE-P       & GAE-P        & Planner       \\
    \midrule
    \multirow{2}{*}{10}
         & $\bar{U}^r$ & $32.08\pm5.35$        & $\mathit{40.26\pm6.13}$   & $\bm{40.38\pm6.13}$  & $33.52\pm5.85$  & $37.52\pm5.45$  & $26.28\pm3.35$  & $38.56\pm6.15$  \\
         & $\bar{E}^r$ & $\mathit{0.015}\,$s   & $>10\,$s                 & $>20\,$s             & $\bm{<0.001}\,$s & $\bm{<0.001}\,$s & $\bm{<0.001}\,$s & $\bm{<0.001}\,$s \\
    \multirow{2}{*}{20}
         & $\bar{U}^r$ & $66.80\pm8.64$        & $77.52\pm8.36$            & $\bm{81.40\pm8.85}$  & $69.53\pm8.26$  & $76.25\pm9.62$  & $63.58\pm8.24$  & $\mathit{78.94\pm9.29}$  \\
         & $\bar{E}^r$ & $\mathit{0.025}\,$s   & $>10\,$s                 & $>20\,$s             & $\bm{<0.001}\,$s & $\bm{<0.001}\,$s & $\bm{<0.001}\,$s & $\bm{<0.001}\,$s \\
    \multirow{2}{*}{30}
         & $\bar{U}^r$ & $104.44\pm10.54$      & $114.82\pm10.10$          & $\bm{124.8\pm11.10}$ & $106.85\pm11.62$& $120.65\pm11.84$& $92.85\pm11.35$ & $\mathit{123.04\pm11.27}$\\
         & $\bar{E}^r$ & $0.040\,$s            & $>12\,$s                 & $>25\,$s             & $\mathit{0.002}\,$s & $\mathit{0.002}\,$s & $\bm{<0.001}\,$s & $\mathit{0.002}\,$s \\
    \multirow{2}{*}{40}
         & $\bar{U}^r$ & $140.78\pm9.81$       & $151.64\pm10.53$          & $\bm{167.72\pm10.16}$& $147.36\pm11.50$& $160.36\pm12.02$& $109.86\pm11.25$& $\mathit{164.46\pm10.32}$\\
         & $\bar{E}^r$ & $0.085\,$s            & $>18\,$s                 & $>31\,$s             & $\mathit{0.002}\,$s & $\mathit{0.002}\,$s & $\bm{<0.001}\,$s & $\mathit{0.002}\,$s \\
    \multirow{2}{*}{50}
         & $\bar{U}^r$ & $176.54\pm16.43$      & $183.28\pm15.68$          & $\bm{209.04\pm16.42}$& $181.25\pm16.25$& $192.95\pm16.72$& $132.86\pm15.02$& $\mathit{206.32\pm16.52}$\\
         & $\bar{E}^r$ & $0.12\,$s             & $>25\,$s                 & $>35\,$s             & $\mathit{0.002}\,$s & $\mathit{0.002}\,$s & $\bm{0.001}\,$s  & $\mathit{0.002}\,$s \\
    \multirow{2}{*}{60}
         & $\bar{U}^r$ & $214.52\pm14.49$      & $222.10\pm14.60$          & $\bm{253.54\pm14.90}$& $224.18\pm15.46$& $243.25\pm15.86$& $165.36\pm15.35$& $\mathit{250.94\pm15.16}$\\
         & $\bar{E}^r$ & $0.15\,$s             & $>30\,$s                 & $>40\,$s             & $\mathit{0.003}\,$s & $\mathit{0.003}\,$s & $\bm{0.001}\,$s  & $\mathit{0.003}\,$s \\
    \multirow{2}{*}{70}
         & $\bar{U}^r$ & $247.56\pm16.04$      & $254.20\pm16.62$          & $\bm{293.94\pm16.22}$& $253.45\pm16.50$& $278.63\pm16.62$& $188.85\pm15.21$& $\mathit{289.20\pm16.26}$\\
         & $\bar{E}^r$ & $0.20\,$s             & $>35\,$s                 & $>50\,$s             & $\mathit{0.003}\,$s & $\mathit{0.003}\,$s & $\bm{0.002}\,$s  & $\mathit{0.003}\,$s \\
    \bottomrule
  \end{tabular}
\end{table*}

\subsection{Dynamic Transfer Stage Experiments} \label{section4.3}

The second study isolates the Transfer policy, whose task is to reallocate Red’s resources on-line in response to Blue’s counter-moves.  Compared with the static pre-allocation stage, this problem is spatio-temporal, coupling local neighbourhood interactions with global network dynamics.

We design baseline algorithms from multiple perspectives — rule-based, optimization-based, and variants removing local/global features — to compare with the Transfer strategy. In particularly, we compare five methods.
Rule-based: Transfer to any neighboring node recaptured by Blue, and hold position, otherwise;
MILP: Solve a per-round MILP to maximize immediate reward (greedy-optimal), ignoring long-term strategy and incurring high computation time;
Global-only: Remove the local GAT module and make decisions rely solely on EGTE global features;
Local-only: Remove EGTE and make decisions solely on local GAT features;
HGformer Transfer (Ours): Full dual-scale agent using EGTE + GAT, trained via PPO. The dynamic stage experiments use the same environment settings as in Section \ref{section4.2}.  Red’s initial deployment is fixed using the trained Planner Agent for consistency. Blue follows a rule-based transfer strategy for all settings. 

Table \ref{transfer} report mean cumulative utility $\bar U^{r}$, transfer cost $\bar C^{r}$, and inference latency $\bar E^{r}$.  HGformer yields the highest utility across all $N$, with especially on $N \ge 40$ graphs.  Rule-based control lacks adaptivity and performs worst.  Per-round MILP is near-optimal per step yet incurs $>100$ s solve time when $N > 30$, preventing full-episode completion.  The Global-only ablation is strategic but imprecise; the Local-only variant is reactive but myopic.  Both trail the dual-scale agent by $4$–$7\%$ utility.  Cost analysis shows that HGformer attains its gains with the lowest $\overline C^{r}$, validating the LFRT feedback that aligns static and dynamic policies. Inference latency remains around 0.2 seconds, confirming its suitability for real-time deployment.

These findings confirm that integrating global and local representations, together with cross-stage feedback, is critical for effective and efficient dynamic reallocation.

\begin{table*}[htbp]
  \centering
  \caption{Performance of different algorithms in the dynamic resource transfer stage .}
  \label{transfer}
  \begin{tabular}{ccrrrrr}
    \toprule
    Size & Metric       & Rule‐based        & MILP                & Local‐only         & Global‐only       & Transfer           \\
    \midrule
    \multirow{3}{*}{10}
      & $\bar U^r$ & $39.92\pm18.80$      & $\mathit{52.92\pm 9.47}$   & $48.12\pm10.13$    & $50.35\pm 9.85$    & $\bm{53.28\pm 8.23}$  \\
      & $\bar C^r$ & $20.20\pm 5.36$      & $\mathit{7.63\pm 4.17}$    & $12.58\pm 8.63$    & $9.32\pm 6.00$     & $\bm{7.38\pm 3.83}$   \\
      & $\bar E^r$ & $0.581\,$s           & $>20\,$s                   & $\bm{0.052}\,$s    & $\mathit{0.063}\,$s & $0.069\,$s           \\
    \midrule
    \multirow{3}{*}{20}
      & $\bar U^r$ & $84.68\pm14.82$      & $\mathit{101.16\pm14.72}$  & $92.36\pm15.98$    & $96.65\pm12.74$    & $\bm{103.52\pm11.36}$ \\
      & $\bar C^r$ & $44.01\pm 4.28$      & $20.03\pm 7.44$            & $36.66\pm 9.78$    & $\mathit{18.63\pm 6.94}$ & $\bm{15.64\pm 6.83}$  \\
      & $\bar E^r$ & $2.675\,$s           & $>50\,$s                   & $\bm{0.078}\,$s    & $\mathit{0.083}\,$s & $0.089\,$s           \\
    \midrule
    \multirow{3}{*}{30}
      & $\bar U^r$ & $122.36\pm25.91$     & $\mathit{152.66\pm16.04}$  & $138.64\pm18.13$   & $146.36\pm15.74$   & $\bm{159.36\pm13.65}$ \\
      & $\bar C^r$ & $65.68\pm 5.56$      & $32.63\pm 8.90$            & $47.36\pm11.25$    & $\mathit{31.98\pm10.78}$ & $\bm{28.68\pm 9.25}$  \\
      & $\bar E^r$ & $3.816\,$s           & $>100\,$s                  & $\bm{0.082}\,$s    & $\mathit{0.095}\,$s & $0.101\,$s           \\
    \midrule
    \multirow{3}{*}{40}
      & $\bar U^r$ & $164.10\pm30.07$     & $\mathit{209.42\pm21.13}$  & $182.32\pm25.45$   & $193.38\pm19.85$   & $\bm{215.45\pm15.96}$ \\
      & $\bar C^r$ & $87.06\pm 5.52$      & $\bm{47.47\pm 9.28}$       & $62.78\pm10.75$    & $56.32\pm 9.65$    & $\mathit{52.77\pm 8.71}$ \\
      & $\bar E^r$ & $5.985\,$s           & $>150\,$s                  & $\bm{0.095}\,$s    & $\mathit{0.101}\,$s & $0.114\,$s           \\
    \midrule
    \multirow{3}{*}{50}
      & $\bar U^r$ & $210.02\pm26.10$     & $\mathit{256.32\pm21.99}$  & $239.34\pm21.36$   & $248.65\pm19.75$   & $\bm{265.88\pm19.62}$ \\
      & $\bar C^r$ & $108.72\pm 6.74$     & $\mathit{60.32\pm10.29}$   & $78.69\pm12.43$    & $70.98\pm11.86$    & $\bm{53.96\pm 7.98}$  \\
      & $\bar E^r$ & $7.872\,$s           & $>250\,$s                  & $\bm{0.105}\,$s    & $\mathit{0.113}\,$s & $0.127\,$s           \\
    \midrule
    \multirow{3}{*}{60}
      & $\bar U^r$ & $247.48\pm26.10$     & $\mathit{313.84\pm23.03}$  & $278.12\pm24.96$   & $293.74\pm26.35$   & $\bm{332.96\pm21.63}$ \\
      & $\bar C^r$ & $129.90\pm 8.23$     & $\mathit{73.36\pm 9.57}$   & $92.34\pm10.25$    & $82.77\pm 9.86$    & $\bm{65.32\pm 8.45}$  \\
      & $\bar E^r$ & $9.689\,$s           & $>350\,$s                  & $\bm{0.114}\,$s    & $\mathit{0.124}\,$s & $0.146\,$s           \\
    \midrule
    \multirow{3}{*}{70}
      & $\bar U^r$ & $281.50\pm39.13$     & $\mathit{362.10\pm28.53}$  & $331.85\pm31.85$   & $357.19\pm29.45$   & $\bm{389.00\pm25.12}$ \\
      & $\bar C^r$ & $153.98\pm 7.18$     & $87.25\pm12.25$            & $102.45\pm16.25$   & $\mathit{83.45\pm18.52}$ & $\bm{78.48\pm10.68}$  \\
      & $\bar E^r$ & $11.132\,$s          & $>500\,$s                  & $\bm{0.123}\,$s    & $\mathit{0.139}\,$s & $0.159\,$s           \\
    \bottomrule
  \end{tabular}
\end{table*}

\begin{table*}[htbp]
  \centering
  \caption{Performance comparison of strategy combinations on the full two-stage game.}
  \label{LFRT}
  \begin{tabular}{ccrrrrrr}
    \toprule
    Size & Metric        & DRule               & DMILP                & GPN‐PPO             & MILP‐T               & HGform($\tilde\theta^P$) & HGform($\hat\theta^P$) \\
    \midrule
    \multirow{3}{*}{10}
      & $\bar U^r$ & $26.18\pm22.73$      & $53.12\pm12.21$       & $42.98\pm15.63$      & $53.05\pm10.27$       & $\mathit{53.28\pm 8.23}$  & $\bm{54.26\pm 6.98}$   \\
      & $\bar C^r$ & $14.71\pm 7.19$      & $ 7.90\pm 3.73$       & $13.36\pm 5.49$      & $ 7.53\pm 4.36$       & $\mathit{7.38\pm 3.83}$   & $\bm{5.93\pm 3.59}$    \\
      & $\bar E^r$ & $1.713\,$s           & $>45\,$s              & $\bm{0.052}\,$s      & $>20\,$s              & $0.069\,$s               & $\mathit{0.068}\,$s     \\
    \midrule
    \multirow{3}{*}{20}
      & $\bar U^r$ & $67.90\pm25.39$      & $101.76\pm15.56$      & $81.69\pm18.28$      & $101.16\pm14.72$      & $\mathit{103.52\pm11.36}$ & $\bm{105.12\pm10.87}$  \\
      & $\bar C^r$ & $40.39\pm 5.62$      & $19.86\pm 6.61$       & $35.77\pm 8.63$      & $20.03\pm 7.44$       & $\mathit{15.64\pm 6.83}$  & $\bm{13.68\pm 5.14}$   \\
      & $\bar E^r$ & $4.188\,$s           & $>90\,$s              & $\bm{0.078}\,$s      & $>20\,$s              & $\mathit{0.089}\,$s      & $\mathit{0.089}\,$s     \\
    \midrule
    \multirow{3}{*}{30}
      & $\bar U^r$ & $108.48\pm27.90$     & $154.56\pm16.95$      & $123.46\pm22.36$     & $155.86\pm17.34$      & $\mathit{159.36\pm13.65}$ & $\bm{161.36\pm10.36}$  \\
      & $\bar C^r$ & $63.28\pm 4.48$      & $34.75\pm 8.48$       & $47.36\pm11.25$      & $\mathit{32.63\pm 8.90}$ & $\mathit{28.68\pm 9.25}$  & $\bm{25.16\pm 8.36}$   \\
      & $\bar E^r$ & $5.696\,$s           & $>120\,$s             & $\bm{0.082}\,$s      & $>25\,$s              & $0.101\,$s               & $\mathit{0.100}\,$s     \\
    \midrule
    \multirow{3}{*}{40}
      & $\bar U^r$ & $153.44\pm35.75$     & $208.68\pm23.35$      & $171.39\pm28.63$     & $209.38\pm21.13$      & $\mathit{215.45\pm15.96}$ & $\bm{216.36\pm11.03}$  \\
      & $\bar C^r$ & $84.60\pm 4.84$      & $47.76\pm10.50$       & $65.98\pm11.53$      & $\mathit{46.12\pm10.02}$ & $52.77\pm 8.71$          & $\bm{40.98\pm 9.32}$   \\
      & $\bar E^r$ & $7.856\,$s           & $>220\,$s             & $\bm{0.095}\,$s      & $>31\,$s              & $\mathit{0.114}\,$s      & $\mathit{0.114}\,$s     \\
    \midrule
    \multirow{3}{*}{50}
      & $\bar U^r$ & $190.74\pm37.71$     & $255.88\pm23.33$      & $231.36\pm20.46$     & $256.32\pm21.99$      & $\mathit{265.88\pm19.62}$ & $\bm{271.36\pm18.85}$  \\
      & $\bar C^r$ & $106.82\pm 6.10$     & $59.81\pm11.24$       & $88.39\pm12.43$      & $61.12\pm13.29$       & $\mathit{53.96\pm 7.98}$  & $\bm{48.25\pm 6.85}$   \\
      & $\bar E^r$ & $9.546\,$s           & $>270\,$s             & $\bm{0.105}\,$s      & $>35\,$s              & $\mathit{0.127}\,$s      & $\mathit{0.127}\,$s     \\
    \midrule
    \multirow{3}{*}{60}
      & $\bar U^r$ & $223.64\pm35.86$     & $315.00\pm22.72$      & $267.12\pm28.16$     & $313.67\pm25.43$      & $\mathit{332.96\pm21.63}$ & $\bm{348.25\pm15.32}$  \\
      & $\bar C^r$ & $128.48\pm 7.13$     & $71.89\pm10.66$       & $102.38\pm10.25$     & $72.16\pm 9.47$       & $\mathit{65.32\pm 8.45}$  & $\bm{50.35\pm 9.85}$   \\
      & $\bar E^r$ & $10.882\,$s          & $>400\,$s             & $\bm{0.114}\,$s      & $>40\,$s              & $\mathit{0.146}\,$s      & $\mathit{0.146}\,$s     \\
    \midrule
    \multirow{3}{*}{70}
      & $\bar U^r$ & $275.26\pm46.47$     & $359.02\pm27.62$      & $328.85\pm32.36$     & $364.51\pm28.53$      & $\mathit{389.00\pm25.12}$ & $\bm{395.71\pm19.35}$  \\
      & $\bar C^r$ & $151.94\pm 7.31$     & $87.79\pm11.23$       & $102.45\pm16.25$     & $86.35\pm12.25$       & $\mathit{78.48\pm10.68}$  & $\bm{70.15\pm 9.35}$   \\
      & $\bar E^r$ & $13.298\,$s          & $>550\,$s             & $\bm{0.123}\,$s      & $>50\,$s              & $\mathit{0.159}\,$s      & $\mathit{0.159}\,$s     \\
    \bottomrule
  \end{tabular}
\end{table*}

\subsection{End-to-End Evaluation}
We now assess HGformer over the complete two-stage Colonel Blotto game, measuring coordinated performance, cost efficiency, and scalability.

We compare HGformer with five representative alternatives. Rule–Rule (DRule): Greedy heuristic for allocation + rule-based transfer; a non-learning baseline; MILP–MILP (DMILP): MILP-based optimization at both stages; an upper-bound reference with high computational cost; GPN–PPO \cite{42}: Decoupled learning via a graph pointer network for allocation and a separate PPO policy for transfer, with no cross-stage feedback; MILP+Transfer(MILP-T): Optimal initial deployment (MILP) combined with our learning-based Transfer Agent; HGformer w/o LFRT(HGform ($\tilde\theta^P$)): Full architecture without feedback—Planner and Transfer trained independently. All methods are evaluated on 100 random graphs for $N\!=\!10$–$70$ under identical hyper-parameters and hardware.  Metrics include cumulative Red reward $\bar U^{r}$, transfer cost $\bar C^{r}$, and mean inference time.

Table \ref{LFRT} summarize the performance of various strategy combinations on the complete two-stage task. HGformer achieves the best overall trade-off across all graph sizes.   On large graphs (e.g., $ N=50$ and $N=70$), HGformer with LFRT surpasses its no-feedback counterpart by more than 8\% in cumulative payoff and reduces transfer cost by approximately 10\%, highlighting the effectiveness of feedback-driven coordination between hierarchical policies. In terms of scalability, DMILP becomes intractable once $N > 30$ because solve times exceed 100 s per round, whereas HGformer sustains 0.2s inference, enabling real-time deployment.  DRule suffers from both low reward and high cost due to its lack of adaptivity, while the partial-learning GPN–PPO cannot capture inter-stage dependencies and therefore underperforms.  Overall, HGformer uniquely combines strategic synergy, cost-aware adaptation, and computational practicality, highlighting its suitability for large-scale, graph-constrained adversarial optimisation.

\section{Concluding Remarks}\label{section5}
This paper investigates the problem of a graph-constrained two-stage dynamic Colonel Blotto game, a setting where traditional methods struggle due to multi-stage coupling and global optimality challenges. We propose HGformer, a hierarchical graph Transformer framework that employs an Enhanced Graph Transformer Encoder (EGTE) to effectively capture global topological and dynamic state features. HGformer integrates an upper-level Planner Agent and a lower-level Transfer Agent, optimized jointly through a Layered Feedback Reinforcement Training (LFRT) mechanism, significantly enhancing strategic coordination and Red’s cumulative payoff.

Experimental results demonstrate HGformer’s superior performance over baseline methods, achieving higher payoffs and lower resource consumption in both initial deployment and dynamic reallocation tasks. Nonetheless, the current framework assumes Blue employs a fixed rule-based policy and that Red has full observability of Blue’s initial actions. Future research will explore adaptive Blue strategies, partial observability scenarios, and equilibrium-based solution concepts to enhance the framework’s robustness and applicability. We also plan to extend HGformer to broader adversarial scenarios, including multi-party and continuous-time resource allocation problems.

\bibliography{reb}

\end{document}